\documentclass{llncs}

\authorrunning{F.G.C. Cabarle, K.C. Bu\~no, H.N. Adorna}
\titlerunning{Time After Time: Notes on Delays In Spiking Neural P Systems}

\usepackage{amsmath,amssymb,latexsym,graphicx,url,algorithm, algorithmic}
\usepackage {graphicx,ulem,textcomp}
\usepackage {todonotes}

\usepackage{epsfig,listings}
\usepackage{hyperref,multirow}
\usepackage[mathcal]{euscript}

\begin{document}

\title{Time After Time: Notes on Delays In Spiking Neural P Systems}
\author{Francis George C. Cabarle, Kelvin C. Bu\~no, Henry N. Adorna}
\institute{
Algorithms \& Complexity Lab\\
Department of Computer Science\\
University of the Philippines Diliman\\
Diliman 1101 Quezon City, Philippines\\
E-mail: {{\tt \{fccabarle,kcbuno,hnadorna\}@up.edu.ph}}}
\maketitle

\begin{abstract}

Spiking Neural P systems, SNP systems for short, are biologically inspired computing devices based on how neurons perform computations. SNP systems use only one type of symbol, the spike, in the computations. Information is encoded in the time differences of spikes or the multiplicity of spikes produced at certain times. SNP systems with delays (associated with rules) and those without delays are two of several Turing complete SNP system variants in literature. In this work we investigate how restricted forms of SNP systems with delays can be simulated by SNP systems without delays. We show the simulations for the following spike routing constructs: sequential, iteration, join, and split.

\end{abstract} 



\section{Introduction}\label{intro-sect}

Membrane computing\footnote{a relatively new field of Unconventional and Natural computing introduced in 1998 by Gheorghe P\u aun, see e.g. \cite{introtomem}} abstracts computational ideas from the way biological cells perform information processing. The models used in Membrane computing are known as P systems. Spiking Neural P systems, or SNP systems for short, are a class of P systems that incorporate the neural-like P systems which only use indistinguishable electric signals or spikes. SNP systems were introduced in \cite{snp} and were proven in \cite{snp} and \cite{universality} to be Turing-complete. Because SNP systems only use one
kind of spike symbol $a$, the output of the computation done by the system is based on the time interval between the first spike and the second spike of a designated output
neuron. This model represents the fact that spikes in a biological neural system are almost identical from an electrical point of view. The time
intervals between these signal spikes are crucial to the computations performed by neurons.

As with the more common results in Membrane computing, SNP systems have been used to efficiently solve hard problems such as the SAT problem using exponential workspace \cite{snpdivision}. Other numerous results also focus on using SNP systems as acceptors, generators, and transducers as in \cite{snp-ibarra} and \cite{snp-alhazov}. SNP systems were also used in order to perform arithmetic operations where the operands are encoded in their binary formats \cite{snp-cpu}.
It has been proven in \cite{snp-ibarra-normal} that delays in applying rules in the neurons of an SNP system are not required for SNP systems to be Turing complete. However SNP systems with delays might still prove to be useful for modeling purposes. In \cite{snp-mat} SNP systems without delays were represented as matrices and the computations starting from an initial configuration can be performed using linear algebra operations. By simulating SNP systems with delays using SNP systems without delays, the work in \cite{snp-mat} can be used to further study SNP systems.\footnote{A massively parallel SNP system simulator in \cite{snpgpu-cmc,snpgpu-romjist} is based on the matrix representation  in order to perform simulations.}

In this work we investigate how SNP systems with delays can be simulated by SNP systems without delays. In particular we focus on a special type of SNP systems that consist of a source neuron and a sink neuron. The initial configuration consists of a spike in the source neuron only and a final configuration where only the sink neuron has a spike. The spike from the source neuron to the sink neuron is routed through other intermediary neurons using four routing primitives: \textit{sequential, iteration, joins,} and \textit{splits}.

The contents of this paper are organized as follows: Section \ref{prelims-sect} provides definitions and notations that will be used in this work. Section
\ref{struct-behav-snp-pn-sect} presents our results. We end this paper in Section \ref{final-remarks-sect} with some final notes, conjectures and open problems.

\section{Preliminaries}\label{prelims-sect}

It is assumed that the readers are familiar with the basics of Membrane Computing \footnote{a good introduction is 
\cite{introtomem}
 with recent results and information in the P systems webpage {at {\tt http://ppage.psystems.eu/}} and a recent handbook in \cite{mem-handbook} }
and formal language theory.
 We only briefly mention notions and notations which will be useful throughout the paper. 
Let $V$ be an alphabet, $V^*$ is the free monoid over $V$ with respect to concatenation and the identity element $\lambda$ (the empty string). The set of all non-empty strings over $V$ is denoted as $V^+$ so $V^+ = V^* - \{\lambda\}$. We call $V$ a \textit{singleton} if $V = \{a\}$ and simply write $a^*$ and $a^+$ instead of $\{a^*\}$ and $\{a^+\}$. The length of a string $w \in V^*$ is denoted by $|w|$. If $a$ is a symbol in $V$, $a^0 = \lambda$.
A language $L \subseteq V^*$ is regular if there is a regular expression $E$ over $V$ such that $L(E) = L$. A regular expression over an alphabet $V$ is constructed starting from $\lambda$ and the symbols of $V$ using the operations union, concatenation, and $+$, using parentheses when necessary to specify the order of operations. Specifically, $(i)$ $\lambda$ and each $a \in V$ are regular expressions, $(ii)$ if $E_1$ and $E_2$ are regular expressions over $V$ then $(E_1 \cup E_2)$, $E_1E_2$, and $E_1^+$ are regular expressions over $V$, and $(iii)$ nothing else is a regular expression over $V$. With each expression $E$ we associate a language $L(E)$ defined in the following way: (i) $L(\lambda) = \{\lambda\}$ and $L(a) = \{a\}$ for all $a \in V$, (ii) $L(E_1 \cup E_2) = L(E_1) \cup L(E_2)$, $L(E_1E_2) = L(E_1)L(E_2)$, and $L(E_1^+) = L(E_1)^+$, for all regular expressions $E_1$, $E_2$ over $V$. Unnecessary parentheses are omitted when writing regular expressions, and $E^+ \cup \{\lambda\}$ is written as $E^*$. 

We define an SNP system of a {finite} degree $m \geq 1$  as follows:

			$$\Pi=(O,\sigma_1,\ldots, \sigma_m, syn),$$
where:
\begin{enumerate}
\item[1.] $O=\{a\}$ is the singleton alphabet ($a$ is called \textit{spike}).

\item[2.] $\sigma_1,\ldots, \sigma_m$ are neurons of the form  $\sigma_{i}=(\alpha_i, {R_i}),1\leq i \leq m$, where:
\begin{enumerate}
{
	\item[(a)] $\alpha_i \geq 0$ is an integer representing the initial number of spikes in $\sigma_i$}
	\item[(b)] ${R_i}$ is a finite set of rules of the general form 
	
	$$E/a^c \rightarrow a^b;d$$
	
	where $E$ is a regular expression over $O$, $c \geq 1$, $b \geq 0$, with $c \geq b$.

\end{enumerate}

	\item[3.] $syn \subseteq \{1, 2, \ldots, m\} \times \{ 1, 2, \ldots, m \}$, $(i,i) \notin syn$ for $1 \leq i \leq m$, are synapses between neurons.

\end{enumerate}

A \textit{spiking rule} is where $b \geq 1$. A \textit{forgetting rule} is a rule where $b = 0$ is written as $E/a^c \rightarrow \lambda$. If $L(E) = \{a^c\}$ then spiking and forgetting rules are simply written as $a^c \rightarrow a^b$ and $a^c \rightarrow \lambda$, respectively. Applications of rules are as follows: if neuron $\sigma_i$ contains $k$ spikes, $a^k \in L(E)$ and $k \geq c$, then the rule $E/a^c \rightarrow a^b \in R_i$ is enabled and the rule can be fired or applied. If $b \geq 1$, the application of this rule removes $c$ spikes from $\sigma_i$, so that only $k - c$ spikes remain in $\sigma_i$. The neuron fires $b$ number of spikes to every $\sigma_j$ such that $(i,j) \in syn$. If the delay $d = 0$, the $b$ number of spikes are sent immediately i.e. in the same time step as the application of the rule. If $d \geq 1$ and the rule with delay was applied at time $t$, then the spikes are sent at time $t+d$. From time $t$ to $t+d-1$ the neuron is said to be \textit{closed}\footnote{this corresponds to the \textit{refractory period} of the neuron in biology} and cannot receive spikes. Any spikes sent to the neuron when the neuron is closed are \textit{lost} or removed from the system. At time $t+d$ the neuron becomes \textit{open} and can then receive spikes again. The neuron can then apply another rule at time $t+d+1$.  If $b = 0$ then no spikes are produced. SNP systems assume a global clock, so the application of rules and the sending of spikes by neurons are all synchronized. 


A configuration of the system can be denoted as $\langle n_1/t_1$, $\ldots$, $n_m/t_m \rangle$, where each element of the vector is the configuration of a neuron $\sigma_i$, with $n_i$ spikes and is open after $t_i \geq 0$ steps. At time step 0, since no rules, with or without delay, are yet to be applied, $t_i = 0, \forall i \in \{1,...,m\}$. $t_i$ increases when $\sigma_i$ applies a rule with delay.

To illustrate, let us have an snp system with three neurons, $\sigma_1$, $\sigma_2$, and $\sigma_3$ with synapses $(1,2)$, $(2,3)$. Only $\sigma_1$ has some number of spikes, say $x$. $\sigma_1$ has a rule $a^+/a \rightarrow a; 2$, $\sigma_2$ has a rule $a \rightarrow a$, and $\sigma_3$ is a sink neuron. At a time step $k$ where no rules are yet to be applied, the configuration of this system would be $\langle x/0$, $0/0$, $0/0 \rangle$. At time step $k+1$, we apply the rule in $\sigma_1$; the configuration would then become $\langle x-1/2,\ 0/0,\ 0/0 \rangle$. At time step $k+2$, $\langle x-1/1,\ 0/0,\ 0/0 \rangle$. At time step $k+3$, $\sigma_1$ is open again and will now fire one spike to $\sigma_2$, the configuration would now be $\langle x-1/0,\ 1/0,\ 0/0 \rangle$. At time step $k+4$, $\sigma_1$ can again use its rule, and now $\sigma_2$ can also apply its rule; the configuration would then become $\langle x-2/2,\ 0/0,\ 1/0 \rangle$.

The initial number of spikes of $\sigma_i$ is denoted as $\alpha_i$, $\alpha_i \geq 0$. The initial configuration therefore is $\langle \alpha_1/0, \ldots, \alpha_m/0 \rangle $.  A \textit{computation} is a sequence of transitions from an initial configuration. A computation may halt (no more rules can be applied for a given configuration) or not. 

\begin{figure}[tb]
	\centering
	\includegraphics[scale=.33]{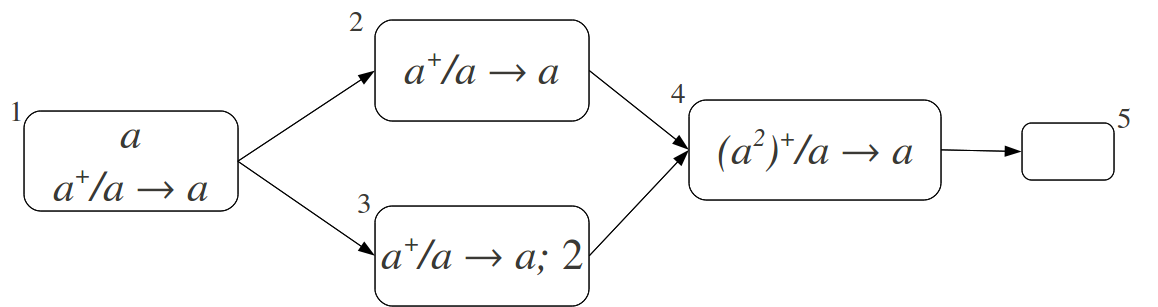} 
	\caption{An example of an SNP system where the source neuron is $\sigma_1$ and sink neuron is $\sigma_5$. Neuron $\sigma_4$ will only spike once it accumulates two spikes, one each from $\sigma_2$ and $\sigma_3$.}
	\label{snp-example-fig}
\end{figure}

For SNP systems in this work\footnote{more information on SNP systems having input and output neurons as well as nondeterministic SNP systems (we focus on deterministic systems in this work) can be found in \cite{snprecent} and \cite{mem-handbook} for example.},we have only one \textit{source} neuron (a neuron without any incoming synapses to it) and one \textit{sink} neuron (a neuron without any outgoing synapses from it). The initial configuration will always be a single spike found only at the source neuron. The objective is to route this spike from the source to the sink neuron. The routes will involve paths in the system, where a \textit{path} consists of at least two neurons $\sigma_i, \sigma_j$ such that $(i,j) \in syn$. Using this idea of a path, we can have four basic routing constructs: (1) sequential, where $\sigma_{i+1}$ spikes after $\sigma_i$ and $(i,i+1) \in syn$, (2) iteration, where at least two neurons spike multiple (possibly an infinite) number of times and a loop is formed e.g. synapses $(i,j)$ and $(j,i)$ exist between neurons $\sigma_i$ and $\sigma_j$, (3) join, where spikes from at least two input neurons $\sigma_m, \sigma_n$ are sent to a neuron $\sigma_i$, where $(m, i), (n, i) \in syn$, so that $\sigma_i$ produces a spike to at least one output neuron $\sigma_j$, and $(i, j) \in syn$, (4) split, where a spike from $\sigma_j$ is sent to at least two output neurons $\sigma_k$ and $\sigma_l$ and $(j,k),(j,l) \in syn$. An example of an SNP system that we deal with in this work is shown in Fig. \ref{snp-example-fig}.

SNP systems considered in this work are those with neurons having exactly one rule only. For SNP systems with delays, notice that if there exists a sequential path from $\sigma_i$ (with delay $d1$) to $\sigma_j$ (with delay $d2$) if $d1 < d2$ and $\sigma_i$ spikes more than once, it is possible for the spikes from $\sigma_i$ to be lost. This is due to the possibility that $\sigma_j$ may still be closed when spikes from $\sigma_i$ arrive. We therefore avoid lost spikes by assuming $d1 \geq d2$ whenever a sequential path from $\sigma_i$ to $\sigma_j$ exists, and leave the cases where $d1 < d2$ as an open problem. Given an SNP system with delays $\Pi$ and an SNP system without delay $ \overline{\Pi}$, by simulation in this work we mean the following: (a) the arrival time $t$ of spikes arrive at sink neurons of $\Pi$ is the the same time $t$ for the arrival of spikes at the sink neurons of $ \overline{\Pi}$ or offset by some constant integer $k$, (b) the number of spikes that arrive in the sink neurons of $\Pi$ is the same for the sink neurons in $ \overline{\Pi}$ or a factor of the delay of $\Pi$. We define the \textit{total runtime} of $\Pi$ and $ \overline{\Pi}$ as the total time needed for a spike to arrive to the sink neurons from the source neurons.

\section{Main Results}\label{struct-behav-snp-pn-sect}

Now we present our results in simulating SNP systems with delays using SNP systems without delays. Once again we denote SNP systems with delays as $\Pi$ and those without delays as $ \overline{\Pi}$, subject to the restrictions and assumptions mentioned in the previous section.

\begin{definition}
Given an SNP system, $\Pi =(O,\sigma_1,\ldots, \sigma_m, syn)$, a routing of $\Pi$, $\Pi'$ is defined as

$$\Pi' = (O, \Sigma', syn'),$$

where:

\begin{enumerate}

	\item[1.] $syn'$ $\subseteq$ $syn$ 
	
	\item[2.] $\Sigma$ is a subset of the set of neurons of $\Pi$. $\Sigma' = \{\sigma, \hat\sigma|(\sigma, \hat\sigma) \in syn'\}$.
	
\end{enumerate}

\end{definition}

%
%
\begin{lemma}

Given an SNP with delay, $\Pi$ that contains a sequential routing, there exists an SNP without delay, $\overline{\Pi}$ that contains a routing that can simulate the sequential routing of $\Pi$.

\end{lemma}

\begin{figure}[tb]
	\centering
	\includegraphics[scale=.33]{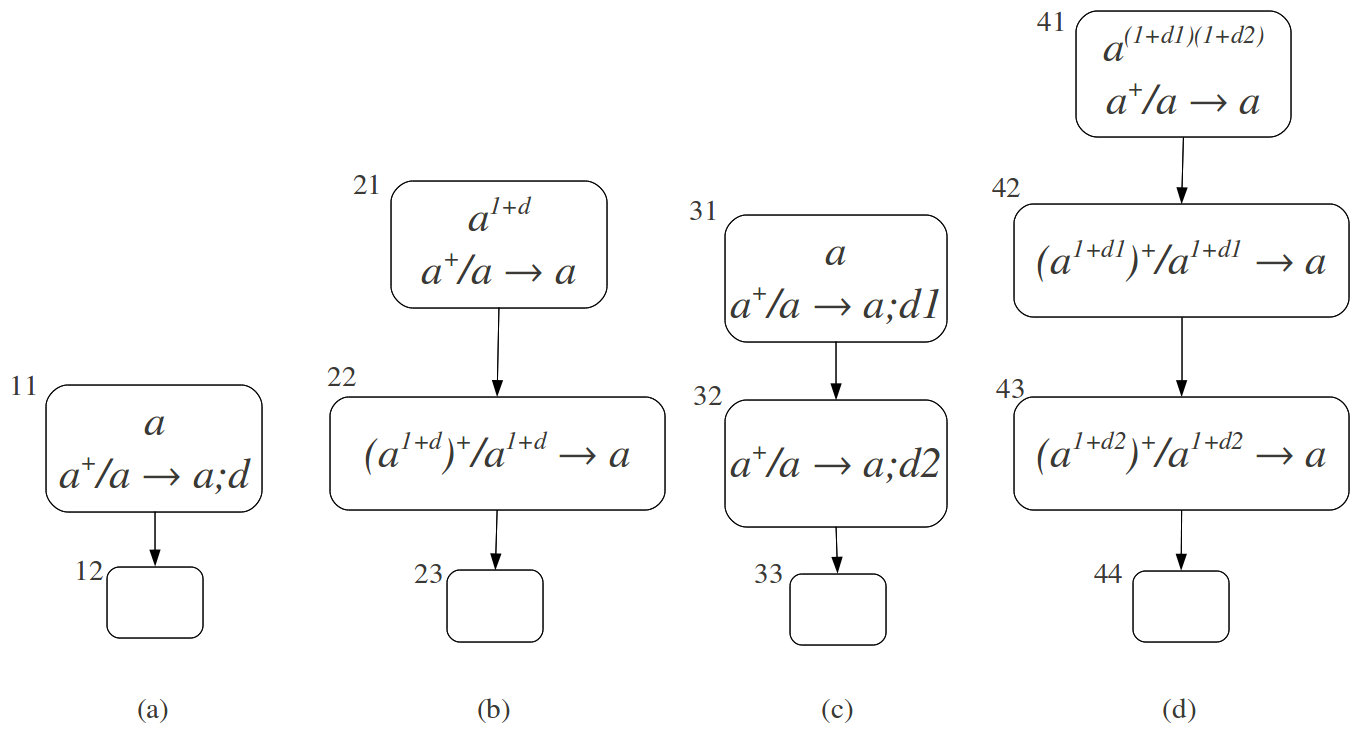} 
	\caption{Sequential routing, with single and multiple delays.}
	\label{seq-one-two-delay-fig}
\end{figure}

\proof We refer to Fig. \ref{seq-one-two-delay-fig} for illustrations and designate empty neurons as sink neurons. Let $\Pi$ be the system in Fig. \ref{seq-one-two-delay-fig}(a). Total runtime for $\Pi$ is $d$ steps, where final spike count for $\sigma_{12}$ is $\alpha_{12} = 1$. Now let $ \overline{\Pi}$ be Fig. \ref{seq-one-two-delay-fig}(b) that simulates $\Pi$. The initial spike count for $\sigma_{21}$ is $ \alpha_{21} = {1+d}$. Total runtime for $ \overline{\Pi}$ is $1 + d$ steps, and $\sigma_{22}$ receives $1+d$ number of spikes. Final spike count for $\sigma_{23}$ is $ \alpha_{23} = 1$.

\begin{figure}[tb]
	\centering
	\includegraphics[scale=.35]{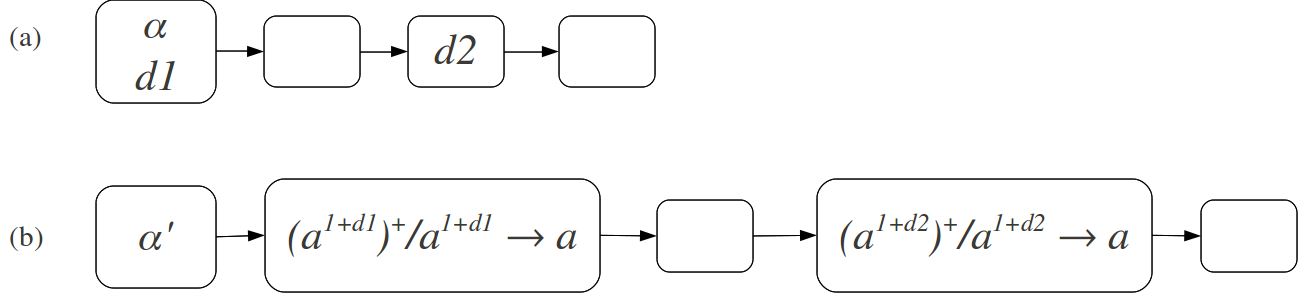} 
	\caption{Sequential routing with additional neurons without delays.}
	\label{seq-two-neur-delay-one-nodelay-fig}
\end{figure}

Now let $\Pi$ be the system in Fig. \ref{seq-one-two-delay-fig}(c) and $ \overline{\Pi}$ be the system in Fig. \ref{seq-one-two-delay-fig}(d). $\Pi$ now has two delays, $d1$ and $d2$, and the total runtime for $\Pi$ is $1 + d1 + d2$. For $ \overline{\Pi}$ (with initial spikes $ \alpha = {(1+d1)(1+d2)}$), the total runtime is $(1+d1)(1+d2) + 3 -1$. The always open neurons of $ \overline{\Pi}$ except for $\sigma_{44}$ each introduce one time delay i.e. 3, minus one time delay since $\sigma_{44}$ does not spike. The final spike count for $\sigma_{33}$ is $\alpha = 1$, which is also the final spike count for $\sigma_{44}$. It can be easily shown that addition of neurons without delayed rules in either $\Pi$ or $ \overline{\Pi}$ contribute to a delay of one time step (per neuron) being added to the total runtimes for both systems (see Fig. \ref{seq-two-neur-delay-one-nodelay-fig}).  \qed

The total runtime for $ \overline{\Pi}$ is one additional time step from the total runtime of $\Pi$. The additional time step for $ \overline{\Pi}$ comes from delay due to the first spiking of $\sigma_{21}$.

%
%
\begin{lemma}

Given an SNP with delay, $\Pi$ that contains an iterative routing, there exists an snp without delay, $\overline{\Pi}$ that contains a routing that can simulate the iterative routing of $\Pi$.

\end{lemma}

\begin{figure}[tb]
	\centering
	\includegraphics[scale=.35]{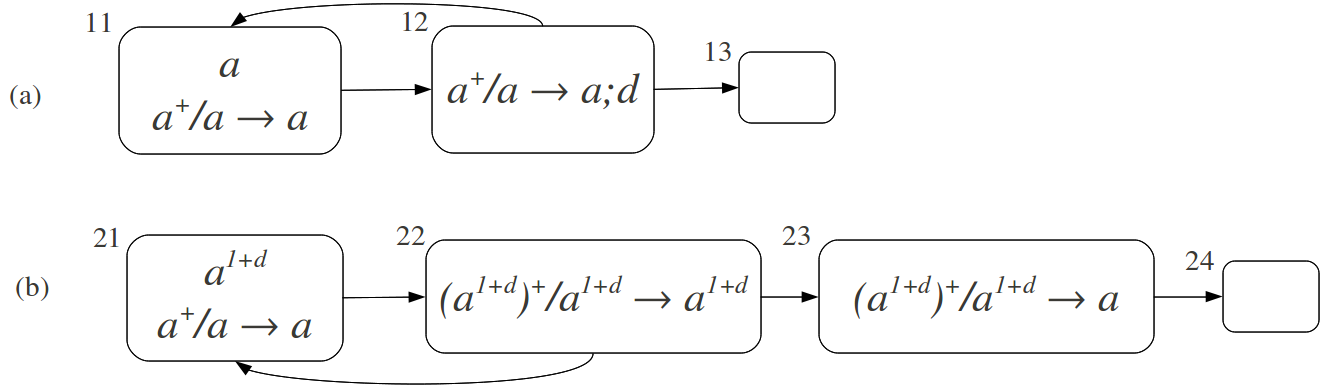} 
	\caption{Iterative routing. Note that the $ \overline{\Pi}$ in (b) that simulates $\Pi$ in (a) is still the same even if $\sigma_{11}$ has the delay instead of $\sigma_{12}$}
	\label{iter-one-delay-fig}
\end{figure}

\begin{figure}[tb]
	\centering
	\includegraphics[scale=.35]{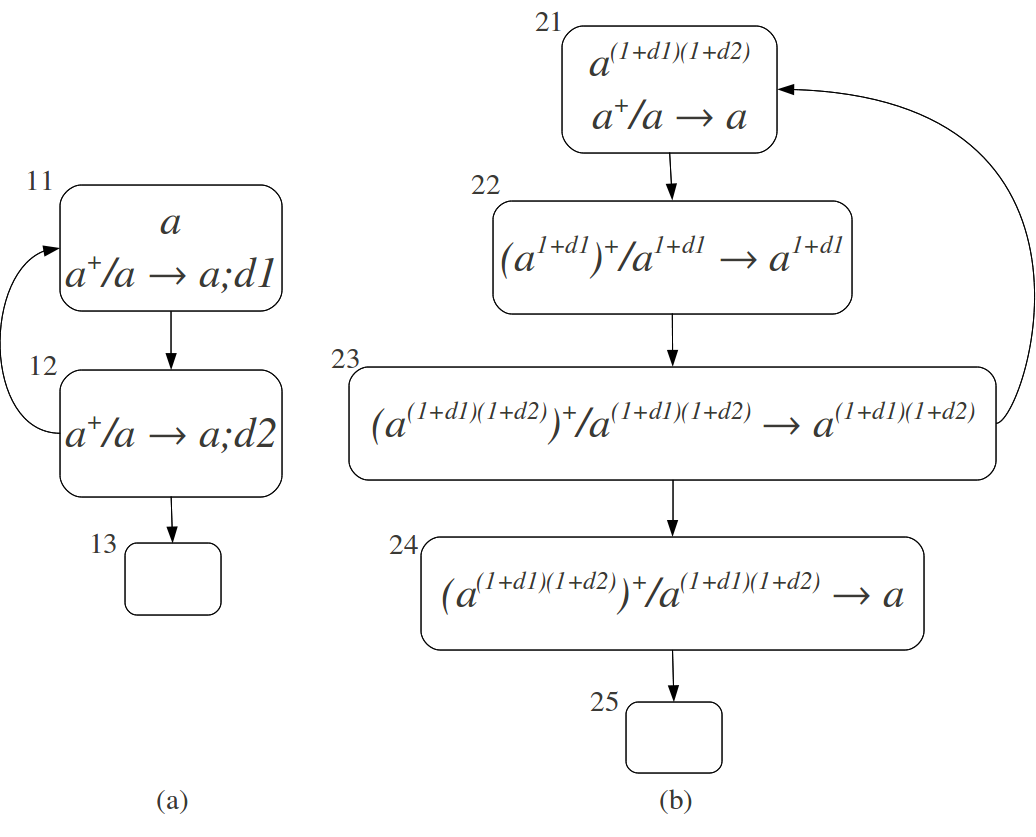} 
	\caption{Iteration with more than one delay.}
	\label{iter-two-delays-fig}
\end{figure}

\proof We refer to Fig. \ref{iter-one-delay-fig} and \ref{iter-two-delays-fig} for illustrations. Let $\Pi$ be the system in Fig. \ref{iter-one-delay-fig}(a) and $ \overline{\Pi}$ be the system in Fig. \ref{iter-one-delay-fig}(b). Notice that $\sigma_{11}$ and $\sigma_{12}$ continuously replenish the spike of one another, forming an infinite loop so that $\sigma_{13}$ will keep on accumulating spikes. The total runtime is $1+d$ steps. Neurons $\sigma_{21}$ and $\sigma_{22}$ similarly replenish the spike of one another in an infinite loop. The total runtime for $ \overline{\Pi}$ is $2 + d$, with one additional delay due to $\sigma_{23}$. Due to the operation of $\sigma_{23}$, both $\sigma_{13}$ and $\sigma_{24}$ obtain one final spike each. It can be easily shown that if the delay of $\Pi$ is at $\sigma_{11}$ instead of $\sigma_{12}$, the $ \overline{\Pi}$ in Fig. \ref{iter-one-delay-fig}(b) can still simulate $\Pi$.

For iterations with more than one delay, let $\Pi$ be the system in Fig. \ref{iter-two-delays-fig}(a) and $ \overline{\Pi}$ be the system in Fig. \ref{iter-two-delays-fig}(b). The total runtime of $\Pi$ is $d1 + d2$ and $(1+d1)(1+d2) + 3 -1$ for $ \overline{\Pi}$. The final spike count for $\Pi$ is $\alpha_{13} = 1$ which is the same as the final spike count for $ \overline{\Pi}$ (due to the operation of $\sigma_{24}$). \qed

Notice that the total runtime for $ \overline{\Pi}$ for iterative routing with multiple delays is the same as the total runtime for $ \overline{\Pi}$ for sequential routing with multiple delays. Again, every additional neuron without delay adds one time delay to the total runtime of $\Pi$ and $ \overline{\Pi}$.

%
%
\begin{lemma}\label{splits-lemma}

Given an SNP with delay, $\Pi$ that contains a split routing, there exists an SNP without delay, $\overline{\Pi}$ that contains a routing that can simulate the split routing of $\Pi$.

\end{lemma}

\begin{figure}[tb]
	\centering
	\includegraphics[scale=.35]{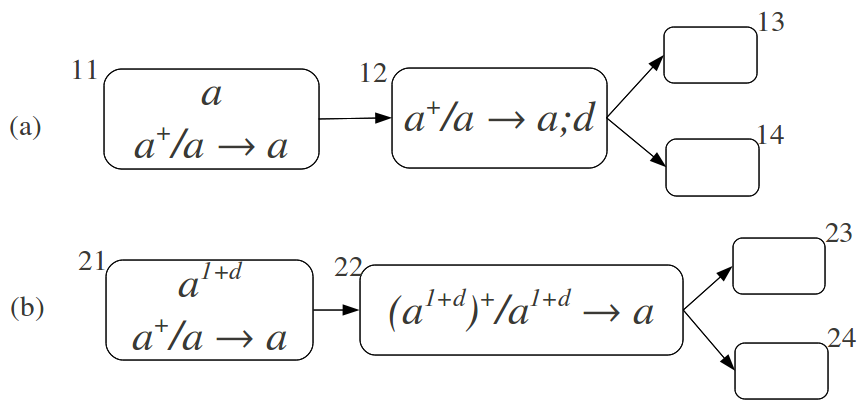} 
	\caption{A split where the neuron performing the split has a delay.}
	\label{split-parent-delay-fig}
\end{figure}

\begin{figure}[tb]
	\centering
	\includegraphics[scale=.35]{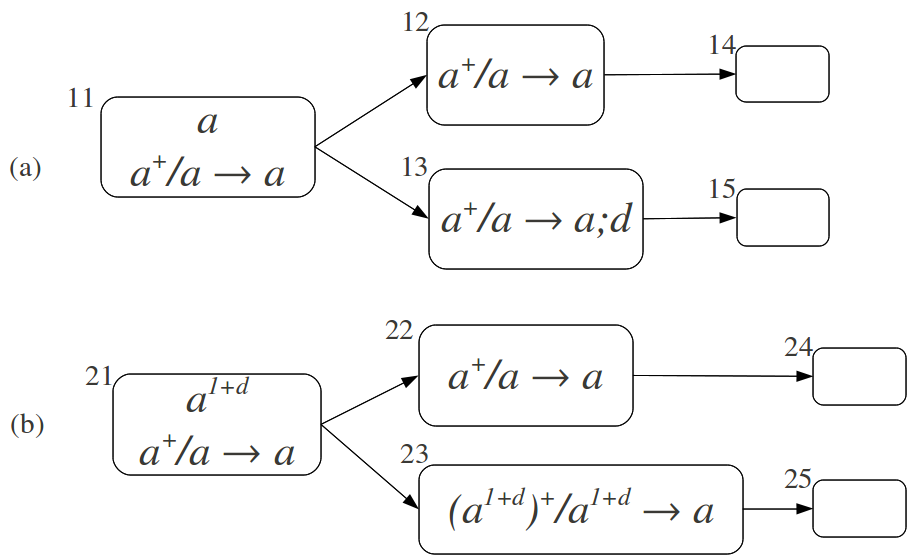} 
	\caption{A split where a child neuron has a delay.}
	\label{split-child-delay-fig}
\end{figure}

\proof We refer to Fig. \ref{split-parent-delay-fig} and \ref{split-child-delay-fig} for illustrations. Let $\Pi$ be the system in Fig. \ref{split-parent-delay-fig}(a) and $ \overline{\Pi}$ be the system in Fig. \ref{split-parent-delay-fig}(b) and we first consider a split where the neuron $\sigma_{12}$ that performs the split is the one that has a delay. The total runtime for $\Pi$ is $1+d$, with a final spike count of $\alpha_{13} = \alpha_{14} = 1$. The total runtime for $ \overline{\Pi}$ is also $1+d$ with a similar final spike count of $\alpha_{23} = \alpha_{24} = 1$. 

Now let $\Pi$ be the system in Fig. \ref{split-child-delay-fig}(a) and $ \overline{\Pi}$ be the system in Fig. \ref{split-child-delay-fig}(b) and we consider a split where the delay is in one of the receiving child neurons ($\sigma_{13}$).  In this case the sink neurons do not receive a spike at the same time. For $\Pi$, if $\sigma_{11}$ spikes at time $t$, $\sigma_{14}$ receives a spike at time $t+1$ while $\sigma_{15}$ receives a spike at time $t+1 + d$. The total runtime therefore (time when both sink neurons obtain one spike) is $1 + d$. For $ \overline{\Pi}$ if $\sigma_{21}$ first spikes at time $t$ then $\sigma_{24}$ first receives a spike at $t+1$. The total runtime is $1+d$ where $\alpha_{25} = 1$ and $\alpha_{24} = 1+d$.  \qed

Note that since $\sigma_{24}$ accumulates $1+d$ spikes (from $\sigma_{21}$), if the split SNP system $ \overline{\Pi}$ is a subsystem of a larger SNP system, an additional neuron $\sigma_i$ with a rule $a^{1+d} \rightarrow a$ needs to be added in order to return the spike number back to one\footnote{this additional neuron once again adds one time step of delay.}

%
%
\begin{lemma}

Given an SNP with delay, $\Pi$ that contains a join routing, there exists an SNP without delay, $\overline{\Pi}$ that contains a routing that can simulate the join routing of $\Pi$.

\end{lemma}

\begin{figure}[tb]
	\centering
	\includegraphics[scale=.35]{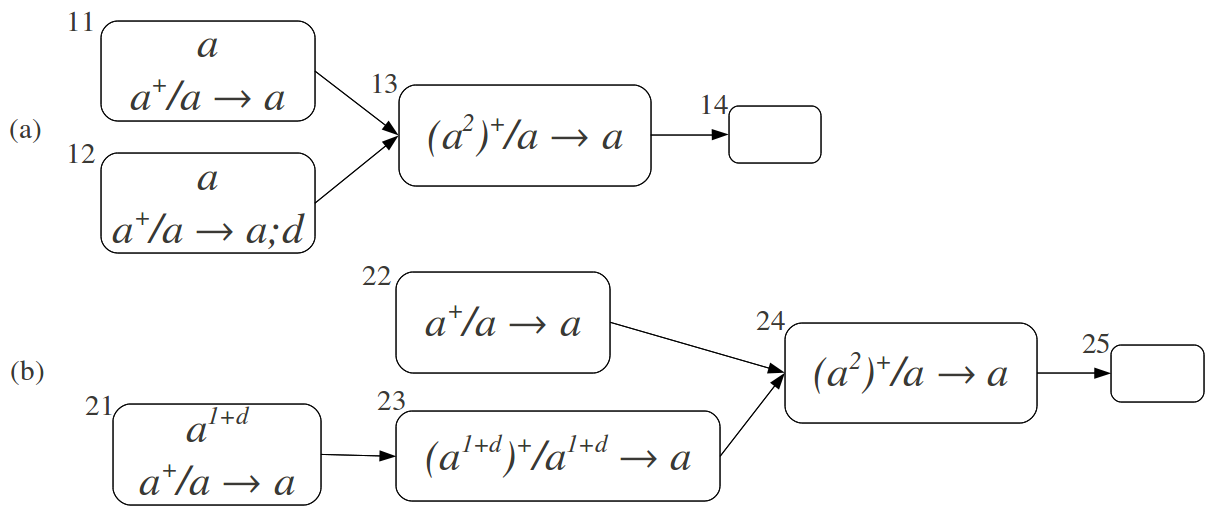} 
	\caption{A join where a parent neuron has a delay.}
	\label{join-parent-delay-fig}
\end{figure}

\begin{figure}[tb]
	\centering
	\includegraphics[scale=.35]{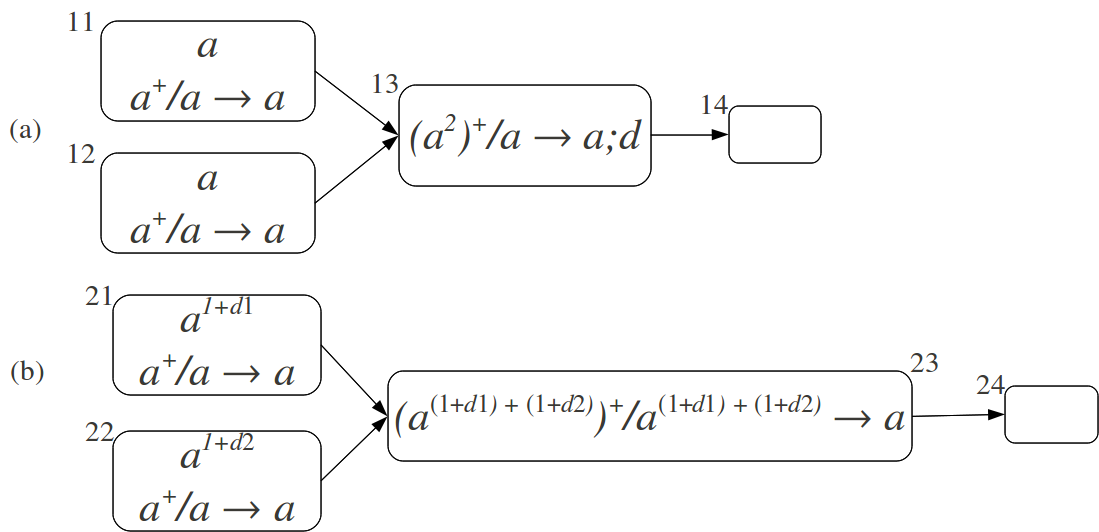} 
	\caption{A join where the neuron performing the join has a delay.}
	\label{join-child-delay-fig}
\end{figure}

\proof We refer to Fig. \ref{join-parent-delay-fig} and \ref{join-child-delay-fig} for illustrations. Let $\Pi$ be the system in Fig. \ref{join-parent-delay-fig}(a) and $ \overline{\Pi}$ be the system in Fig. \ref{join-parent-delay-fig}(b) so that we first we consider a join where the delay is in a neuron $\sigma_{12}$ before the neuron $\sigma_{13}$ that performs the join.  The total runtime for $\Pi$ is $1 +d$ since $\sigma_{13}$ will wait for the spike from parent neuron $\sigma_{12}$ to arrive before sending a spike to $\sigma_{14}$. The total runtime for $ \overline{\Pi}$ is $2 + d$ because of the additional source neuron $\sigma_{21}$, since $\sigma_{24}$ will wait for the spike from $\sigma_{23}$. For $\Pi$ and $ \overline{\Pi}$ the final spike count is 1.

Now let $\Pi$ be the system in Fig. \ref{join-child-delay-fig}(a) and $ \overline{\Pi}$ be the system in Fig. \ref{join-child-delay-fig}(b) and we consider next a join where the delay is in the neuron $\sigma_{13}$ that performs the join. Total runtime for $\Pi$ is $1 + d$ which is also the total runtime for $ \overline{\Pi}$. The final spike count for both systems is 1.  \qed

\begin{theorem}
\label{thm-summary}

Let $\Pi$ be an SNP system with delays that follows the restrictions and assumptions given in Section \ref{prelims-sect} and contains routings limited to the following: 

\begin{enumerate}

	\item sequential
	
	\item iterative
	
	\item split
	
	\item join
	
\end{enumerate}

Then, there exists an SNP system without delay, $ \overline{\Pi}$, that simulates $\Pi$.

\end{theorem}

\proof Proof follows from Lemma 1, 2, 3, and 4. \qed

\section{Final Remarks}\label{final-remarks-sect}

Since $ \overline{\Pi}$ has to simulate $\Pi$, it is expected that either additional neurons, regular expressions, or spikes are to be added to $ \overline{\Pi}$ in order to simulate the delay(s) and final configuration of $\Pi$ (with some offset or spike count multiple of $\Pi$). We have shown in this work how to the routing constructs of $\Pi$, although we could perhaps do better i.e. less neurons, less initial spikes, and so on.

\begin{conjecture}
For the case $\alpha = a^k$ where $k > 1$ is an integer, SNP systems without delays can still simulate SNP systems with delays.
\end{conjecture}

\begin{conjecture}
It is possible to make the time of spiking of SNP systems with delays exactly coincide with the time of spiking of SNP systems without delays (or at least to lessen the neurons or time difference as presented in this work).
\end{conjecture}

We did not consider a split where the delays of the child neurons are not equal, since Lemma \ref{splits-lemma} only considers a split where only one of the child neurons have a delay (also, it is easy to show that if both child neurons of a split have the same delay $d$ then the $ \overline{\Pi}$ in Lemma \ref{splits-lemma} still holds). As mentioned earlier, we also leave as an open problem on how to simulate SNP systems with delays where $d1 < d2$ for sequential routing.

\section*{Acknowledgments}
F.G.C. Cabarle is supported by the {DOST-ERDT program} and the UP Information Technology Dev't Center. K.C. Bu\~ no is supported by the UP Diliman Department of Computer Science (UPD DCS). H.N. Adorna is funded by a {DOST-ERDT research grant} and the Alexan professorial chair of the {UPD DCS}. 


\end{document}